\documentclass{article}

\usepackage{microtype}
\usepackage{graphicx}
\usepackage{subfigure}
\usepackage{booktabs} 
\usepackage{gensymb}
\usepackage{hyperref}

\usepackage[accepted]{icml2019}

\icmltitlerunning{VolMap: A Real-time Model for Semantic Segmentation of LiDAR surrounding view}

\begin{document}

\twocolumn[
\icmltitle{VolMap: A Real-time Model for Semantic \\ Segmentation of a LiDAR 360\degree{} surrounding view}




\begin{icmlauthorlist}
\icmlauthor{Hager Radi}{h}
\icmlauthor{Waleed Ali}{h}
\end{icmlauthorlist}

\icmlaffiliation{h}{Department of CDV-DSF, Valeo Interbranch Automotive Software, Cairo, Egypt}

\icmlcorrespondingauthor{Hager Radi}{hagerradi@aucegypt.edu}
\icmlcorrespondingauthor{Waleed Ali}{waleed.ali@valeo.com}

\icmlkeywords{LiDAR, semantic segmentation, surrounding view, 360 LiDAR, Machine Learning, 3D, ICML}

\vskip 0.3in
]



\printAffiliationsAndNotice{}  

\begin{abstract}
This paper introduces VolMap, a real-time approach for the semantic segmentation of a 3D LiDAR surrounding view system in autonomous vehicles. We designed an optimized deep convolution neural network that can accurately segment the point cloud produced by a 360\degree{} LiDAR setup, where the input consists of a volumetric bird-eye view with LiDAR height layers used as input channels. We further investigated the usage of multi-LiDAR setup and its effect on the performance of the semantic segmentation task. Our evaluations are carried out on a large scale 3D object detection benchmark containing a LiDAR cocoon setup, along with KITTI dataset, where the per-point segmentation labels are derived from 3D bounding boxes. We show that VolMap achieved an excellent balance between high accuracy and real-time running on CPU.
\end{abstract}

\section{Introduction} \label{sec:intro}
 \begin{figure}[ht]
\vskip 0.2in
\begin{center}
\centerline{\includegraphics[width=0.7\columnwidth, height=7cm]{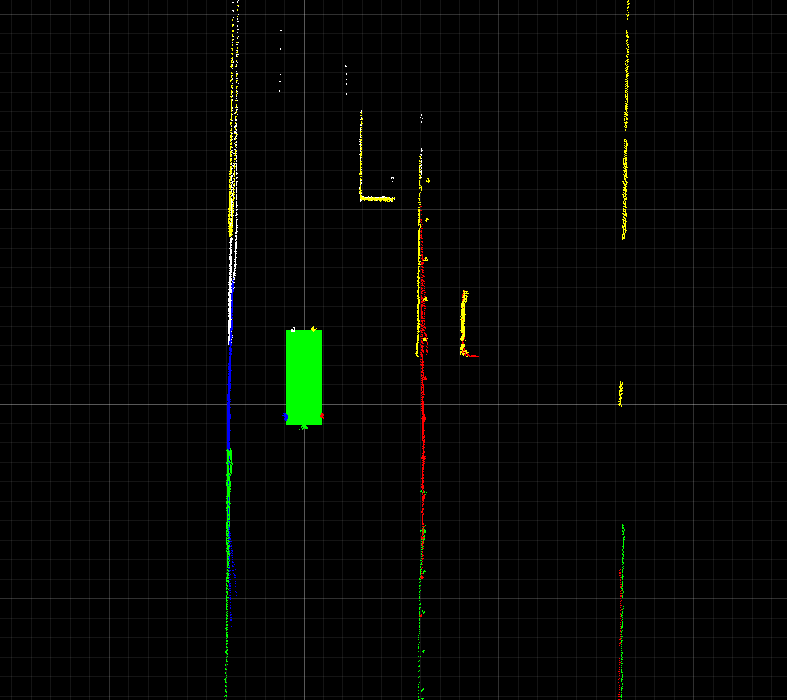}}
\caption{Sample of a LiDAR surrounding view point cloud; color code represents the reflections unique to each sensor.}
\label{fig:sensor_setup1}
\end{center}
\vskip -0.2in
\end{figure}
Autonomous driving systems require an accurate classification and localization of the perceived road objects. Such systems depend on different sensors like camera, LiDAR, RADAR and Ultrasonic sensors. LiDAR sensors play an important role for environment perception in self-driving cars as they can provide a dense 3D mapping of the surrounding environment. This work is focused towards a specific LiDAR setup, which is the 360\degree{} surrounding view system, sometimes referred to as LiDAR cocoon.\\
Semantic Segmentation task can be defined as a point-wise classification for the input point cloud. This task is considered as a part of the global task of scene understanding in which visual information has to be associated with an entity while considering the spatial information. Many approaches tackled the problem of semantic segmentation on LiDAR point clouds. A few traditional methods focused on understanding the semantic features of the 3D shapes. With the rise of deep learning research, Convolution Neural networks have been used in a 3D manner to operate on the point cloud modeled as volumes. The volumetric CNNs achieved good progress towards solving the segmentation task but the sparsity of these volumes and the associated unnecessary computation remained a problem given the high cost of 3D CNNs.\\
A few later works introduced the usage of view-based projections to reduce the run-time. Front view or bird-eye view projections produced good accuracy in 3D object detection as seen in MV3D \cite{mv3d}. Then, Spherical \cite{squeezeseg} Projections were proposed to score well in terms of run-time and accuracy. Among all of the approaches mentioned above, there is always a trade-off between coming up with acceptable accuracy and running in real-time. Our contribution lies in:
 \begin{itemize}
\item Achieving Real-time performance for a 3D LiDAR surrounding view segmentation on CPU.
\item Exploiting a lightweight 2D segmentation model for a 3D input point cloud.
\item Using LiDAR Layers as input channels for the segmentation network
\item Enabling accurate semantic segmentation for a multi-LiDAR surrounding view system
\end{itemize}

The paper is structured as follows: Section 2 reviews all the related work in the literature. Section 3 introduces the problem and the motivation behind this work while section 4 describes the approach, network structure and the data used. In Section 5, experimentation results are reported and compared against other state-of-the-art architectures. Finally, an ablation study is conducted to investigate the effect of multi-LiDAR setup on the segmentation task.


\section{Related Work} \label{sec:lit_review}
In this section, a comprehensive literature review is conducted for the semantic segmentation task in 3D point clouds.\\
\begin{figure}[ht]
\vskip 0.2in
\begin{center}
\centerline{\includegraphics[width=\columnwidth, height=10cm]{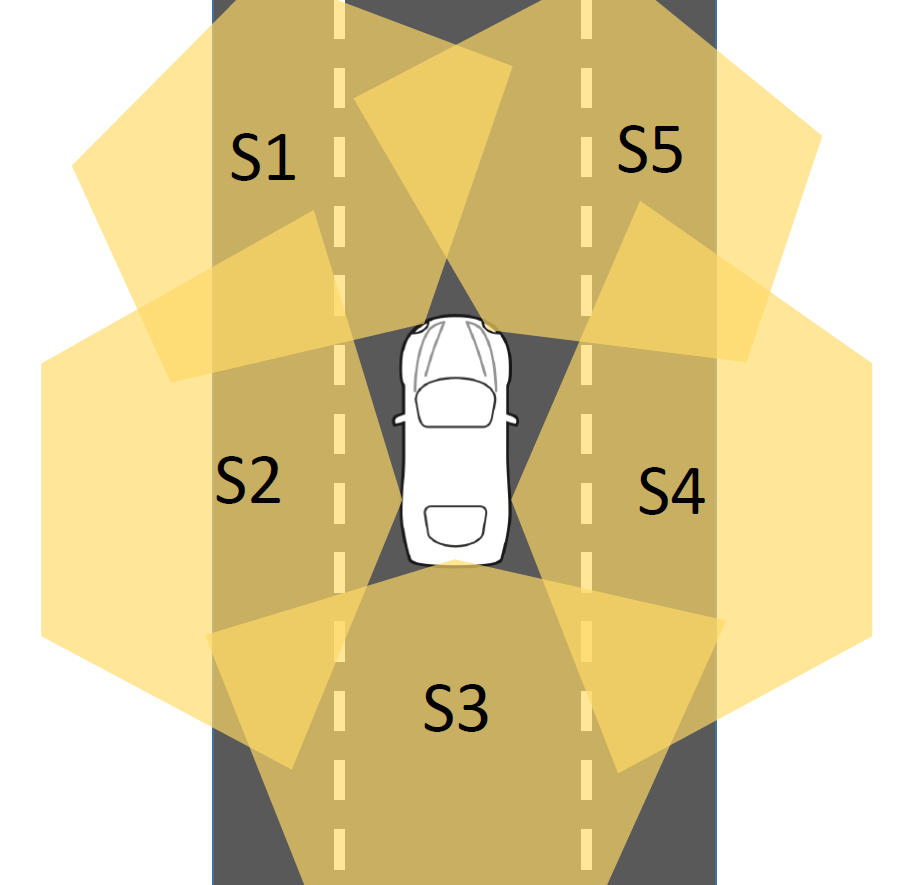}}
\caption{LiDAR surrounding view Sensor setup}
\label{fig:sensor_setup}
\end{center}
\vskip -0.2in
\end{figure}

\textbf{Feature Engineering} based methods usually depend on different stages such as ground removal, clustering and attribute calculation based on traditional methods as in \cite{douillard2011} that involved ground removal or in \cite{moosmann2009} that performed neighborhood graph construction followed by postprocessing and geometrical calculations. In \cite{dube2018}, dynamic oxeloxel grids are used in which the point normals are updated selectively to cache the geometric consistencies to an incremental segmentation algorithm. In addition, Voting for Voting \cite{voting2015} proposed using a 3D sliding window similar to convolution that was tested on KITTI \cite{kitti} object detection Benchmark.\\
\textbf{Volumetric CNNs} or Voxelization is another approach widely used in the literature. VoxNet \cite{voxnet2015} created a volumetric Occupancy Grid representation with a supervised 3D Convolutional Neural Network. In 3D shapeNets \cite{3dshapenets2015}, a convolutional deep belief network was proposed to represent a geometric 3D shape on a 3D voxel grid to recognize and reconstruct objects from a single-view 2.5D depth map. Bo Li \cite{li20173d} trained a 3D FCN to detect vehicles on KITTI \cite{kitti} Lidar data. Charles et. al \cite{volumetric2016} proposed a 3D CNN trained along with an image-based 2D Network In Network(NIN) to classify the 2D projection of the original 3D shape. Multi-view CNN \cite{multiview2015} introduced the usage of different views parsed with a CNN to extract view-based features followed by a view-pooling to obtain the overall classification. Most of these approaches are computationally-expensive due to using 3D convolutional layers and operating on (around 90\%) sparse voxel grid cells. \\
\textbf{Unordered Point Clouds} has gained more attention in the past few years to learn directly from the raw data. PointNet \cite{pointnet} and PointNet++ \cite{pointnet++} were one of the top papers that initiated this approach by applying NN models on raw 3D point clouds directly without the need to transform into volumes. A few follow-up papers worked on capturing geometric details from the raw data. PointCNN \cite{pointcnn} introduced an x-conv layer instead of a multi-layer perceptron(MLP) to aggregate information from neighbourhoods. Dynamic Graph CNN \cite{dynamicgraph} proposed a grouping method to replace the one in PointNet++ by constructing a local neighborhood graph and applying convolution-like operations on the edges. Although this approach was a success in classifying indoor shapes, it was not extensively tested on outdoor scenes specifically in the autonomous driving context.\\
\textbf{View-based Projections} have been proposed to project 3D point clouds into multiple views where 2D operations are applied. This approach improves the run-time for 3D point cloud inputs. MV3D \cite{mv3d} fused the Lidar front view and bird-eye view along with the RGB image to generate 3D boxes on KITTI \cite{kitti}. As a result of this fusion, it outperformed the fully 3D approaches in terms of both run-time and accuracy. SqueezeSeg \cite{squeezeseg} proposed spherical projections for Lidar data to construct a 2D image using the intensity, range, x, y and z coordinates as input channels; SqueezeNet is employed as its backbone network with a conditional random field as a post-processing step. PointSeg \cite{pointseg} extended SqueezeSeg model to emphasize on the efficiency in run-time and accuracy. All of these approaches achieved efficient inference time on LiDAR data maintaining good accuracy.

\section{Problem} \label{sec:problem}
\begin{figure}[ht]
\vskip 0.2in
\begin{center}
\begin{tabular}{@{}c@{}}
    \centerline{\includegraphics[width=\columnwidth,height=45pt]{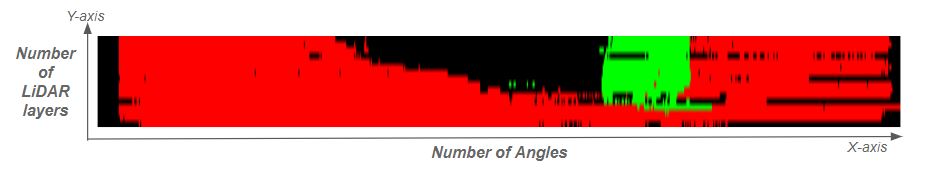}}
  \end{tabular}
  \caption{Spherical Projection view of LiDAR point cloud}\label{fig:pgm}
  \end{center}
\vskip -0.2in
\end{figure}

Our main goal is to perform point-wise classification on the point cloud resulting from a LiDAR 360\degree{} setup in one shot. The system designed is a LiDAR surrounding view consisting of 5 sensors, mounted in the setup shown in Figure \ref{fig:sensor_setup}. In our literature review, the state-of-the-art approaches have been discussed while some of them were closely investigated due to their reported high performance. PointNet++ \cite{pointnet++} processes unordered point clouds directly so the input is not affected by the number of LiDAR sensors or the overall surrounding view. It can better capture the geometry of irregular shapes in the point clouds while objects do not occlude each other. However, the model's run-time is computationally expensive, which is not tolerated in real-time applications on embedded devices like autonomous driving.\\
View-based approaches, especially the spherical projections \cite{squeezeseg}, have proved to achieve good performance for point cloud segmentation that can run in real-time. In Figure \ref{fig:pgm}, a LiDAR point cloud is shown after being translated to the spherical projection representation. The image height represents the number of layers in a LiDAR sensor, and the image width represents the number of unique angles at which points are reflected. This approach is based on the fact that LiDAR beams detect the frontal objects occluding all what is behind.\\

\begin{figure}[ht]
\vskip 0.2in
\begin{center}
    \centering
    \centerline{\includegraphics[width=\columnwidth, height=100pt]{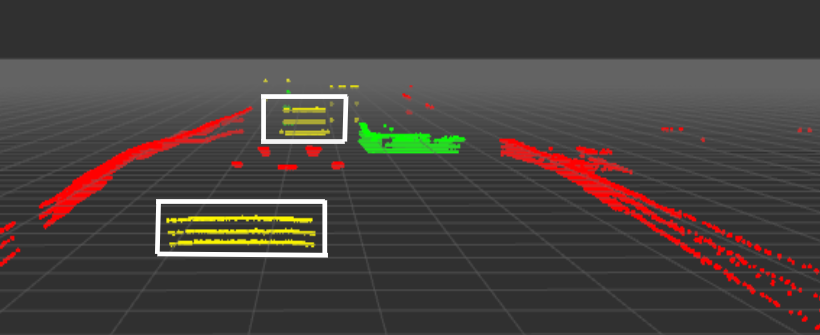}}
    \caption{Occulusion case in multi-LiDAR point cloud}
    \label{fig:pgm_failed}
    \end{center}
\vskip -0.2in
\end{figure}

For our LiDAR surrounding view setup, all the five sensors are projected to a single point of reference resulting in s 3D point cloud that may have points from different objects occluding each other, as seen in Figure \ref{fig:pgm_failed}. Hence, the resulting LiDAR point cloud cannot be represented as a sphere which makes the spherical projection \cite{squeezeseg} not applicable for the multi-LiDAR surrounding view setup.\\

To overcome the issue mentioned above, it is a need to come up with a real-time approach that maintains a good accuracy on point-wise classification. Inspired by the Voxelization in VoxelNet \cite{voxelnet}, the same approach was followed to represent a LiDAR point cloud as a voxel Grid map, where points on both x and y are discretized with a certain resolution as done in bird-eye view projection while LiDAR layers are represented in the height as they are. This produces a volumetric bird-eye view shape maintaining all the height information in order not to lose those features. The number of layers in LiDAR can be known in advance and so does the height as the third dimension. This volumetric map is further used as an input to a 2D segmentation network, where the third dimension(layers) is used as input channels.\\
Key differences distinguish VolMap from VoxelNet \cite{voxelnet} as the latter (1) discretizes the z dimension (height) the same way as in x and y, (2) uses 3D convolutions right after the input layers which adds expensive computational cost.

\begin{figure}[ht]
\vskip 0.2in
\begin{center}
    \begin{tabular}{@{}c@{}}
    \centerline{\includegraphics[width=\columnwidth,height=120pt]{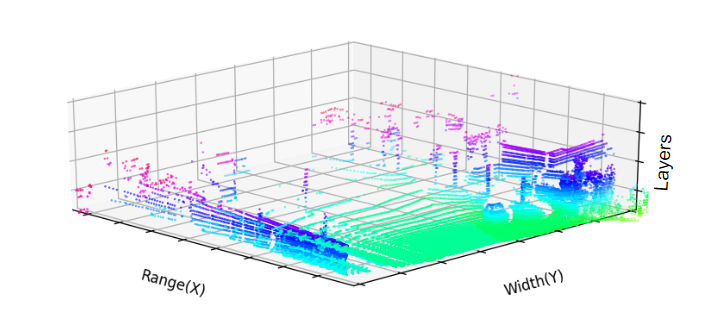}}
  \end{tabular}
  \caption{VolMap input representation of a LiDAR point cloud}\label{fig:volmap}
  \end{center}
\vskip -0.2in
\end{figure}
\section{Method Description} \label{sec:approach}
\subsection{Datasets} \label{subsec:data}
Two main LiDAR datasets have been used in this paper. Since a 360\degree{} LiDAR point cloud is targeted, SCALA, Valeo's scanning laser, has been used along with its internal dataset in our experiments. The dataset consists of both highway and urban scenes with Car and Truck annotated classes. The annotations are provided as 3D bounding boxes which were used to extract the point-wise segmentation labels.\\
KITTI \cite{kitti} is a public dataset for autonomous driving tasks; however, there is no benchmark for semantic segmentation so the 3D object detection benchmark has been used to extract the point-wise classifications as mentioned earlier in SCALA.
\subsection{Input Representation} \label{sec:input}
LiDAR sensors produce a point cloud in which all the surrounding objects are detected in the 3D world. Figure \ref{fig:pgm_failed} shows a sample of a 3D LiDAR point cloud. A volume is constructed around this cloud as seen in Figure \ref{fig:volmap} where its the first dimension is the range(x), the second dimension is the width(y) and the third dimension is the number of LiDAR layers. As a result, the intensity of the existing points in each cell is used to indicate the occupancy. \\
A region of interest is assumed around the vehicle in the four directions, in order to calculate the shape of the volume. A certain resolution is used in both the X and Y directions, where each cell in X and Y are represented with a vector of size N, equal to the number of LiDAR layers to preserve the height information.\\
This yields a volume of \textbf{height}: Range(X) / resolution(x) , and \textbf{width}: Range(Y) / resolution(y) and \textbf{depth} equals the number of layers.
\subsection{Network Structure} \label{subsec:nn}
\begin{figure}[ht]
\vskip 0.2in
\begin{center}
\centerline{\includegraphics[width=1.1\columnwidth, height=200pt]{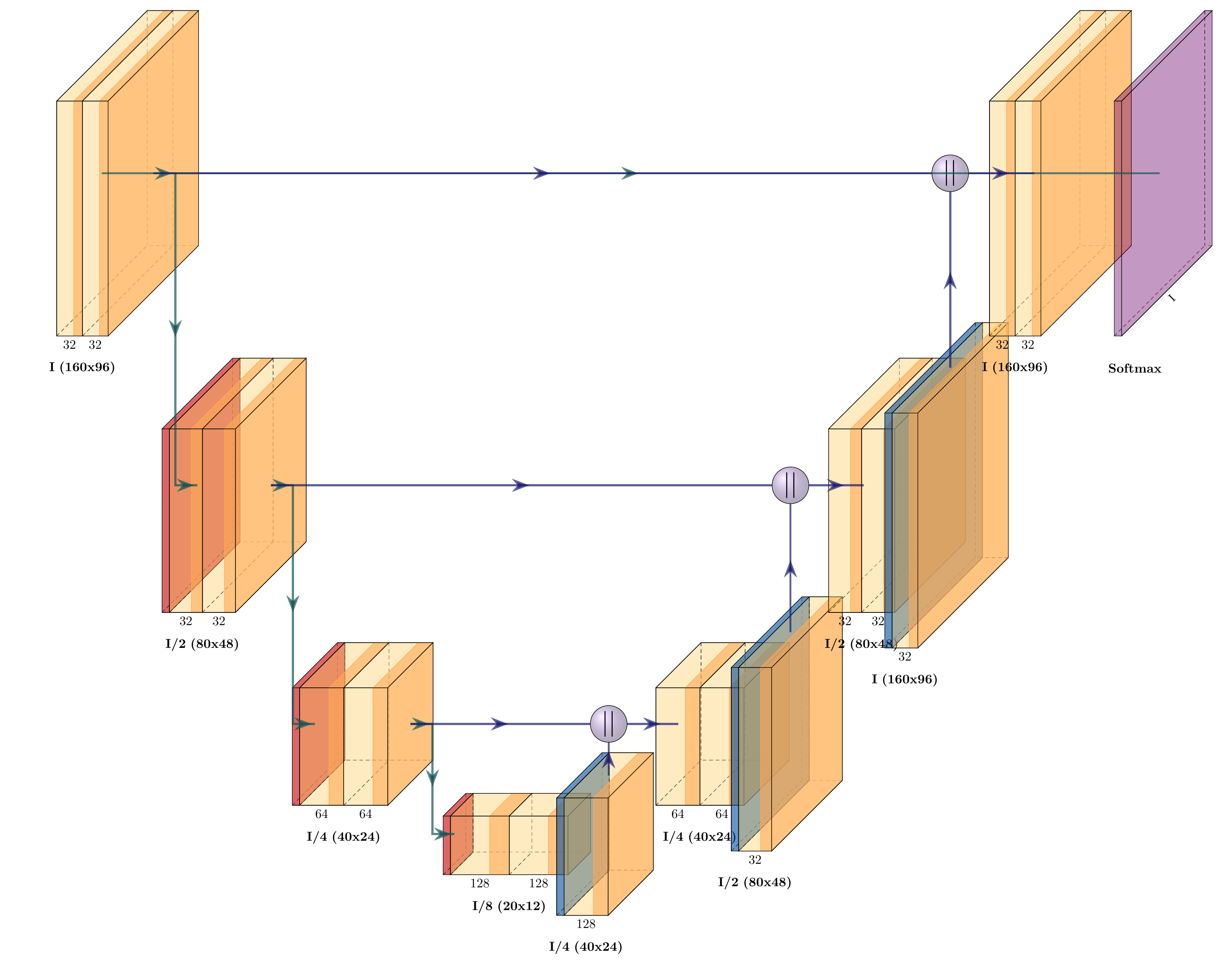}}
\caption{VolMap - a lightweight version of Unet for semantic Segmentation of a Lidar surrounding view}
\label{fig:volMap_arch}
\end{center}
\vskip -0.2in
\end{figure}

Our goal is to perform semantic segmentation on a 360\degree{} point cloud. VolMap, a lightweight version of Unet \cite{unet} has been used as in the architecture shown in Figure \ref{fig:volMap_arch}. VolMap model differs from the original Unet in the number of feature channels in the convolution layers, and the usage of upsampling layers instead of deconvolution. Added to that, only three levels of downsampling are used since the nature of our data does not require such a deep network. 

\textbf{The loss}: Weighted cross entropy loss is used, as described in enet \cite{enet}, where weights are assigned based on the frequency of the classes encountered, following this equation:
\begin{equation}
\label{loss_eq}
loss = \frac{1.0} {log(1.02 + class\_frequency)}
\end{equation}
\textbf{The metric}: The Intersection Over Union (IOU) is calculated for each class to evaluate our models. For the results reported in section \ref{sec:exp_results}, the test data is projected back to the raw point cloud to avoid data loss; hence, the IOU is measured on the original point cloud without any data transformations.

\section{Experiments} \label{sec:exp_results}
VolMap is evaluated on two dataset benchmarks: SCALA, and KITTI \cite{kitti} as described in section \ref{subsec:data}. Models have been trained for 300 epochs, with the architecture described in section \ref{subsec:nn} using stochastic gradient descent (SGD), where a learning rate of 1e-4 and a batch size of 16 point clouds were applied.
\subsection{Evaluation on SCALA benchmark}
\subsubsection{Implementation details}
To evaluate our model’s performance, two main models are used to compare against: SqueezeSeg \cite{squeezeseg} as a real-time view-based approach and PointNet++ \cite{pointnet++} as an unordered point-cloud based approach. \\
\textbf{SqueezeSeg}: As discussed in section \ref{sec:problem}, a LiDAR surrounding view point cloud cannot be modeled with a spherical projection. As an alternative, points from each sensor are modeled in the spherical projection format. The input data coming from the five sensors have been augmented and trained altogether. This model suffered from over-fitting when trained this way given the loss noticed in the input features of each sensor to the network. A projection of each sensor's point cloud was not able to capture enough features for our classes.\\
\textbf{PointNet++}: The original model from the paper \cite{pointnet++} was used to train the 360\degree{} point cloud as raw data coming directly from the sensors so that the input shape is the number of points detected in each time stamp.\\
\textbf{VolMap}: The input representation described in subsection \ref{sec:input} has been followed to construct our input data. In SCALA 360\degree{} surrounding system experiments, the chosen Region Of Interest goes to 30 meters in both the front and back directions(60m range) and 20 meters in both the right and left directions(40m width). The resolution used in both X and Y is 0.4 meters, which gives us a volume of [160, 98, 4] used as the input to our segmentation network. Different resolutions and regions of interest have been investigated to reach the chosen ones. Given the IOU results on the raw data, the current configurations are the best fit for both our run-time and accuracy.
\subsubsection{Results}
Evaluation has been done on a sampled subset from the original dataset with sampling rate of 100 frames per second on completely different scenarios than the ones included in the training/validation. The three models are trained and compared using the same data splits for training, validation and testing. VolMap is able to achieve higher accuracies than PointNet++ while being 10x faster in the run-time as shown in tables \ref{table_1} and \ref{table_2}. VolMap can capture the geometry of the irregular shapes found in SCALA point cloud even with its high sparsity, while PointNet++ requires dense input data to be able to excel the same task.
\begin{table}[t]
\caption{Run-time results on SCALA 360\degree{} Surrounding View}
\label{table_1}
\begin{center}
\begin{small}
\begin{sc}
\begin{tabular}{lcccr}
\toprule
Model & SqueezeSeg  & PointNet++ & VolMap \\
  \midrule
Run-time & N/A & 200ms & \textbf{19.8}ms \\
\bottomrule
\end{tabular}
\end{sc}
\end{small}
\end{center}
\end{table}
\begin{table}
 \caption{IOU results on SCALA 360\degree{} Surrounding View}
 \label{table_2}
 \vskip 0.15in
   \begin{center}
  \begin{small}
\begin{sc}
 \begin{tabular}{lccr}
   \toprule
    Model & SqueezeSeg & PointNet++ & VolMap \\
   \midrule
    CAR & N/A & 42\% & \textbf{85.3}\% \\ 
    Truck & N/A & 54\% &\textbf{64.9}\% \\ 
   \bottomrule
 \end{tabular}
 \end{sc}
\end{small}
\end{center}
\vskip -0.1in
\end{table}
\subsection{Evaluation on KITTI benchmark}
\subsubsection{Implementation details}
KITTI \cite{kitti} dataset has been used to train and test a modified implementation of SqueezeSeg \cite{squeezeseg} and PointNet++ \cite{pointnet++} against VolMap to prove the feasibility of our approach.
Although KITTI's LiDAR covers a 360\degree{} view of the scene, only the camera field of view (90\degree{}) has been taken into consideration as this is the only annotated part of the data-set.\\
\textbf{SqueezeSeg-Modified}: The approach introduced in SqueezeSeg \cite{squeezeseg} has been followed with slight changes in the input shape. SqueezeSeg uses the range, intensity and the three Cartesian coordinates of each point as input channels while our implementation uses the range and intensity only. The primary neural network used is the same as the one used in VolMap and described in section \ref{subsec:nn}. The input is constructed as 80 LiDAR layers as image height and 600 angles as image width, while the range and intensity are the two input channels. \\
\textbf{PointNet++}: Original PointNet++ model has been used, while the training followed the self-incremental learning approach as described in \cite{abdou2019End-to-End}. \\
\textbf{VolMap}: To prepare the VolMap input, the 90\degree{} view has been discretized with a resolution of 0.4 meters for the range and width, maintaining ten layers in height used as input channels. This yields an input shape of [160,112,10] to our neural network, which covers a range of 60 meters and a width of 45 meters.
\begin{table}[t]
 \caption{Run-time comparisons on KITTI}
 \label{table_3}
\begin{center}
\begin{small}
\begin{sc}
\begin{tabular}{lccr}
  \toprule
  Model & SqueezeSeg-Mod & PointNet++ & VolMap \\
  \midrule
  Run-time & 45.5ms & 500ms & \textbf{25.7}ms \\
  \bottomrule
 \end{tabular}
\end{sc}
\end{small}
\end{center}
\end{table}
\subsubsection{Results}
Evaluation has been done on data sequences already excluded from the original dataset during training/validation. Tables \ref{table_3} and \ref{table_4} show that VolMap achieves better accuracies on all the classes with \textbf{20x} faster run-time than PointNet++ since it operates on the 3D raw data directly which requires higher inference time than VolMap. As for the Spherical Projecions, VolMap obtained better accuracies on 60\% of the classes with a 50\% faster run-time. This indicates that VolMap can better capture the features of the LiDAR detected objects while maintaining the run-time as an edge over the spherical projection.
\begin{table}[ht]
\caption{IOU results on KITTI}
\label{table_4}
\begin{center}
\begin{small}
\begin{sc}
 \begin{tabular}{lcccccr}
   \toprule
   Model  & SqueezeSeg-Mod & PointNet++ & VolMap \\
   \midrule
    Car & \textbf{84.0\%} & 64.3\% & 79.71\%\\ 
    Van &  52.0\% & 37.61\% & \textbf{60.41\%} \\ 
    Truck & \textbf{70.0\%} & 52.29\% & 67.78\% \\ 
    Pedest. & 26.0\% & 15.33\% & \textbf{33.62\%} \\
    Cyclist & 41.0\% & 22.96\% & \textbf{53.28\%} \\
   \bottomrule
 \end{tabular}
 \end{sc}
\end{small}
\end{center}
\end{table}
\section{Ablation study on LiDAR Surrounding View} \label{sec:ab_study}
Further Experiments have been conducted to study the effect of adding a single LiDAR sensor vs. adding a LiDAR 360\degree{} surrounding view in a vehicle. While KITTI \cite{kitti} datatset provides a 360\degree{} LiDAR point cloud, only the front camera field of view is annotated which is a total of 90\degree{}. That's why KITTI was not part of this study and SCALA benchmark as described earlier in the data section \ref{subsec:data} is used. \\
The LiDAR setup designed in figure \ref{fig:sensor_setup} is the main system under investigation. Different sensor configurations have been used in this experiment, starting from a single frontal left sensor (S1), a single front right sensor (S5), two left and right sensors (S1 and S5) in the front direction to the full 5-sensor setup covering 360\degree{} view, as seen in Figure \ref{fig:sensor_setup}.\\
Results are reported in table \ref{table_5}. Capturing the point cloud structure and geometry improves as more points are detected from different LiDAR sensors. This indicates that the more sensors added in the vehicle, the better understanding VolMap can reach. A sample of the resulting point cloud is shown in Figure \ref{fig:sensor_setup1}; different colors distinguish the reflections coming from each sensor. Multi-LiDAR setup adds more features to each object detected and hence increases the understanding of the overall scene.
\section{Conclusion and Future Work} \label{conclusion}
Pointwise classification for LiDAR point clouds has been a challenging area of research given the non-regular features and the sparsity in the data. Efforts are made to achieve good accuracy to understand the features in unordered point clouds while keeping an eye on the run-time complexities. Most of the current approaches have been tested and tuned for our problem, but none of them achieved the required run-time and accuracy. Using LiDAR Layers/height as input channels to a 2D segmentation network to segment whole 3D surrounding view point cloud was able to get the best balance between real-time running and accurate predictions. \\
The same approach can be extended to 3D object detection, which is an expensive model. Using the LiDAR layers as input channels will replace the 3D Convolutions layers with 2D ones; hence, the same approach will be followed for 3D object detection on KITTI dataset using VoxelNet model architecture. \\
Another problem to investigate is the effect of a single 360 LiDAR vs. multi-LiDAR sensor setup covering the 360\degree{} view on the overall scene understanding.
\begin{table}
\caption{An ablation study on LiDAR surrounding view}
\label{table_5}
\vskip 0.15in
\begin{center}
\begin{small}
\begin{sc}
 \begin{tabular}{lcccccr}
  \toprule
  Setup  & Car & Truck \\
  \midrule
    1 Front Sensor(Right) & 59.21\% & 40.58\%\\  
    1 Front Sensor(Left) & 67.83\% & 23.55\%\\ 
    2 Front Sensors & 71.79\% & 47.18\% \\
    360\degree{} surrounding view setup & 85.37\% & 64.9\%\\ 
  \bottomrule
 \end{tabular}
 \end{sc}
\end{small}
\end{center}
\vskip -0.1in
\end{table}
\section*{Acknowledgements}
We thank Mohamed Zahran for for reviewing this paper and suggesting the ablation study idea presented in section \ref{sec:ab_study}.
\bibliography{volmap}

\begin{thebibliography}{21}
\providecommand{\natexlab}[1]{#1}
\providecommand{\url}[1]{\texttt{#1}}
\expandafter\ifx\csname urlstyle\endcsname\relax
  \providecommand{\doi}[1]{doi: #1}\else
  \providecommand{\doi}{doi: \begingroup \urlstyle{rm}\Url}\fi

\bibitem[Abdou et~al.(2019)Abdou, Elkhateeb, Sobh, and
  El-sallab]{abdou2019End-to-End}
Abdou, M., Elkhateeb, M., Sobh, I., and El-sallab, A.
\newblock End-to-end 3d-pointcloud semantic segmentation for autonomous
  driving.
\newblock 2019.

\bibitem[Chen et~al.(2017)Chen, Ma, Wan, Li, and Xia]{mv3d}
Chen, X., Ma, H., Wan, J., Li, B., and Xia, T.
\newblock Multi-view 3d object detection network for autonomous driving.
\newblock In \emph{Proceedings of the IEEE Conference on Computer Vision and
  Pattern Recognition}, pp.\  1907--1915, 2017.

\bibitem[Douillard et~al.(2011)Douillard, Underwood, Kuntz, Vlaskine, Quadros,
  Morton, and Frenkel]{douillard2011}
Douillard, B., Underwood, J., Kuntz, N., Vlaskine, V., Quadros, A., Morton, P.,
  and Frenkel, A.
\newblock On the segmentation of 3d lidar point clouds.
\newblock In \emph{2011 IEEE International Conference on Robotics and
  Automation}, pp.\  2798--2805. IEEE, 2011.

\bibitem[Dub{\'e} et~al.(2018)Dub{\'e}, Gollub, Sommer, Gilitschenski,
  Siegwart, Cadena, and Nieto]{dube2018}
Dub{\'e}, R., Gollub, M.~G., Sommer, H., Gilitschenski, I., Siegwart, R.,
  Cadena, C., and Nieto, J.
\newblock Incremental-segment-based localization in 3-d point clouds.
\newblock \emph{IEEE Robotics and Automation Letters}, 3\penalty0 (3):\penalty0
  1832--1839, 2018.

\bibitem[Geiger et~al.(2012)Geiger, Lenz, and Urtasun]{kitti}
Geiger, A., Lenz, P., and Urtasun, R.
\newblock Are we ready for autonomous driving? the kitti vision benchmark
  suite.
\newblock In \emph{2012 IEEE Conference on Computer Vision and Pattern
  Recognition}, pp.\  3354--3361. IEEE, 2012.

\bibitem[Li(2017)]{li20173d}
Li, B.
\newblock 3d fully convolutional network for vehicle detection in point cloud.
\newblock In \emph{2017 IEEE/RSJ International Conference on Intelligent Robots
  and Systems (IROS)}, pp.\  1513--1518. IEEE, 2017.

\bibitem[Li et~al.(2018)Li, Bu, Sun, Wu, Di, and Chen]{pointcnn}
Li, Y., Bu, R., Sun, M., Wu, W., Di, X., and Chen, B.
\newblock Pointcnn: Convolution on x-transformed points.
\newblock In \emph{Advances in Neural Information Processing Systems}, pp.\
  828--838, 2018.

\bibitem[Maturana \& Scherer(2015)Maturana and Scherer]{voxnet2015}
Maturana, D. and Scherer, S.
\newblock Voxnet: A 3d convolutional neural network for real-time object
  recognition.
\newblock In \emph{2015 IEEE/RSJ International Conference on Intelligent Robots
  and Systems (IROS)}, pp.\  922--928. IEEE, 2015.

\bibitem[Moosmann et~al.(2009)Moosmann, Pink, and Stiller]{moosmann2009}
Moosmann, F., Pink, O., and Stiller, C.
\newblock Segmentation of 3d lidar data in non-flat urban environments using a
  local convexity criterion.
\newblock In \emph{2009 IEEE Intelligent Vehicles Symposium}, pp.\  215--220.
  IEEE, 2009.

\bibitem[Paszke et~al.(2016)Paszke, Chaurasia, Kim, and Culurciello]{enet}
Paszke, A., Chaurasia, A., Kim, S., and Culurciello, E.
\newblock Enet: A deep neural network architecture for real-time semantic
  segmentation.
\newblock \emph{arXiv preprint arXiv:1606.02147}, 2016.

\bibitem[Qi et~al.(2016)Qi, Su, Nie{\ss}ner, Dai, Yan, and
  Guibas]{volumetric2016}
Qi, C.~R., Su, H., Nie{\ss}ner, M., Dai, A., Yan, M., and Guibas, L.~J.
\newblock Volumetric and multi-view cnns for object classification on 3d data.
\newblock In \emph{Proceedings of the IEEE conference on computer vision and
  pattern recognition}, pp.\  5648--5656, 2016.

\bibitem[Qi et~al.(2017{\natexlab{a}})Qi, Su, Mo, and Guibas]{pointnet}
Qi, C.~R., Su, H., Mo, K., and Guibas, L.~J.
\newblock Pointnet: Deep learning on point sets for 3d classification and
  segmentation.
\newblock In \emph{Proceedings of the IEEE Conference on Computer Vision and
  Pattern Recognition}, pp.\  652--660, 2017{\natexlab{a}}.

\bibitem[Qi et~al.(2017{\natexlab{b}})Qi, Yi, Su, and Guibas]{pointnet++}
Qi, C.~R., Yi, L., Su, H., and Guibas, L.~J.
\newblock Pointnet++: Deep hierarchical feature learning on point sets in a
  metric space.
\newblock In \emph{Advances in Neural Information Processing Systems}, pp.\
  5099--5108, 2017{\natexlab{b}}.

\bibitem[Ronneberger et~al.(2015)Ronneberger, Fischer, and Brox]{unet}
Ronneberger, O., Fischer, P., and Brox, T.
\newblock U-net: Convolutional networks for biomedical image segmentation.
\newblock In \emph{International Conference on Medical image computing and
  computer-assisted intervention}, pp.\  234--241. Springer, 2015.

\bibitem[Su et~al.(2015)Su, Maji, Kalogerakis, and
  Learned-Miller]{multiview2015}
Su, H., Maji, S., Kalogerakis, E., and Learned-Miller, E.
\newblock Multi-view convolutional neural networks for 3d shape recognition.
\newblock In \emph{Proceedings of the IEEE international conference on computer
  vision}, pp.\  945--953, 2015.

\bibitem[Wang \& Posner(2015)Wang and Posner]{voting2015}
Wang, D.~Z. and Posner, I.
\newblock Voting for voting in online point cloud object detection.
\newblock In \emph{Robotics: Science and Systems}, volume~1, pp.\  10--15607,
  2015.

\bibitem[Wang et~al.(2018{\natexlab{a}})Wang, Shi, Yun, Tai, and Liu]{pointseg}
Wang, Y., Shi, T., Yun, P., Tai, L., and Liu, M.
\newblock Pointseg: Real-time semantic segmentation based on 3d lidar point
  cloud.
\newblock \emph{arXiv preprint arXiv:1807.06288}, 2018{\natexlab{a}}.

\bibitem[Wang et~al.(2018{\natexlab{b}})Wang, Sun, Liu, Sarma, Bronstein, and
  Solomon]{dynamicgraph}
Wang, Y., Sun, Y., Liu, Z., Sarma, S.~E., Bronstein, M.~M., and Solomon, J.~M.
\newblock Dynamic graph cnn for learning on point clouds.
\newblock \emph{arXiv preprint arXiv:1801.07829}, 2018{\natexlab{b}}.

\bibitem[Wu et~al.(2018)Wu, Wan, Yue, and Keutzer]{squeezeseg}
Wu, B., Wan, A., Yue, X., and Keutzer, K.
\newblock Squeezeseg: Convolutional neural nets with recurrent crf for
  real-time road-object segmentation from 3d lidar point cloud.
\newblock In \emph{2018 IEEE International Conference on Robotics and
  Automation (ICRA)}, pp.\  1887--1893. IEEE, 2018.

\bibitem[Wu et~al.(2015)Wu, Song, Khosla, Yu, Zhang, Tang, and
  Xiao]{3dshapenets2015}
Wu, Z., Song, S., Khosla, A., Yu, F., Zhang, L., Tang, X., and Xiao, J.
\newblock 3d shapenets: A deep representation for volumetric shapes.
\newblock In \emph{Proceedings of the IEEE conference on computer vision and
  pattern recognition}, pp.\  1912--1920, 2015.

\bibitem[Zhou \& Tuzel(2018)Zhou and Tuzel]{voxelnet}
Zhou, Y. and Tuzel, O.
\newblock Voxelnet: End-to-end learning for point cloud based 3d object
  detection.
\newblock In \emph{Proceedings of the IEEE Conference on Computer Vision and
  Pattern Recognition}, pp.\  4490--4499, 2018.

\end{thebibliography}
\bibliographystyle{icml2019}





\end{document}